# Synthetic-to-Real Imagery Substitution and Augmentation for Multispectral Landmine Detection

**EVAN LI[1], BHARGAV MANDAKOLATHUR[1], ARYAN RUIA[1], KABIR RAGHAVAN[1], JAMES E. GALLAGHER[1], and EDWARD J. OUGHTON[1]**

[1]Department of Geography and Geoinformation Science, George Mason University, Fairfax, VA 22030 USA
Corresponding author: James E. Gallagher (e-mail: jgalla5@gmu.edu).

**ABSTRACT** Landmines caused more than 6,000 recorded casualties in 2024. Of these casualties, 90% were civilians, and nearly half of those civilians were children. Unmanned aerial systems (UAS) equipped with multispectral sensors and computer vision models can accelerate demining operations by detecting surface-laid mines. However, detection accuracy is constrained by scarce labeled landmine training data, which is difficult to collect and rarely matches the environment where models deploy. This research evaluates whether synthetic imagery can close this data gap by testing sixteen training datasets spanning several real-to-synthetic ratios. We trained two computer vision models, YOLOv11 and RF-DETR, on each unique dataset, producing 32 models. We evaluated every model on a test dataset of 1,320 real RGB-LWIR images, covering 264 combinations of season, illumination, altitude, and sensor fusion level. Synthetic imagery came from three sources: Blender scenes constructed using a Vision Language Model (VLM), a Generative Adversarial Network (GAN) applied to real-world imagery, and the same GAN applied to the Blender-VLM renders. Replacing half of the real imagery with Blender renders or GAN-refined real images reduced mean average precision (mAP@0.5) by only 1.7 to 2.6%, whereas the same substitution using GAN-refined Blender-VLM renders reduced performance by 17.3 to 19.9%. No training dataset ratio mix outperformed real-only training. However, augmenting the complete real-world dataset with all synthetic sources increased YOLOv11 from 71.4% to 76.3% mAP, while RF-DETR performed best on real imagery alone. These results demonstrate that synthetic imagery can replace up to half of a real multispectral training dataset with negligible performance loss.

## I. INTRODUCTION

Landmines remain one of the most persistent and devastating post-conflict humanitarian threats, continuing to cause casualties long after hostilities have ended [1], [2]. In 2024, 6,279 people were killed or injured by landmines and explosive remnants of war, the highest annual total since 2020, with 90% of the victims being civilians and approximately half of those civilians being children [2]. As war-torn communities attempt to rebuild, minefields render large tracts of otherwise productive land unusable, and traditional detection and clearance methods remain both inefficient and extremely expensive [3], [4]. Compounding the problem, new mines are emplaced each year in ongoing conflicts such as the Myanmar civil war and the war in Ukraine [5]. As a result, traditional technology struggles to detect mines as quickly as they are laid. For example, clearing Ukraine alone is estimated to take a century and more than $34 billion [5]. The need for faster survey and detection methods is therefore pressing.

As that need has grown, several technologies have emerged that promise more efficient demining. Unmanned aerial systems (UAS) have become increasingly viable platforms for landmine detection, allowing operators to survey minefields without personal risk while covering far larger areas than manual methods [6]. They are also highly adaptable, accommodating a range of sensors including visible-light cameras, thermal imagers, and ground-penetrating radar [1]. In parallel, deep learning-based object detectors have improved rapidly in both accuracy and inference speed, making them increasingly suitable for edge deployment [7], [8]. Among these, the You Only Look Once (YOLO) family has become the dominant approach for real-time detection [9], [10]. Over the last decade, it has evolved into an efficient, modular architecture that consistently outperforms two-stage detectors such as Faster R-CNN in inference speed while remaining competitive in accuracy [11]. Unlike two-stage detectors, which first generate region proposals and then classify them, YOLO frames detection as a single regression problem, predicting bounding boxes and class probabilities simultaneously in one forward pass [8]. Successive releases have also introduced targeted improvements in small-object detection, which is critical for identifying landmines, and in feature extraction, making the family a strong candidate for automated demining workflows [12], [13], [14].

Each sensor modality carried by a UAS has distinct strengths and weaknesses, and neither is sufficient on its own. RGB cameras capture fine spatial detail and distinguish colors and textures, information that is often useful for identifying mine casings [15]. However, they perform poorly in low light or when a mine visually blends into the surrounding surface [16]. Long-wave infrared (LWIR) cameras detect emitted radiance rather than reflected light, making them less sensitive to illumination and allowing mines to be identified by the temperature contrast between the mine casing and the soil [6], [13]. LWIR sensing is nevertheless subject to thermal crossover, in which target and background temperatures converge, reducing contrast and therefore detection accuracy [6]. Multispectral fusion combines the two sensors, retaining the spatial detail of the visible band alongside the thermal contrast and low-light performance of the infrared band [17], [18]. Where one modality suffers, the other one complements [19]. Accordingly, in the dataset used for this study, images are constructed at eleven levels of LWIR contribution, from 0% LWIR (fully visible) to 100% LWIR, in 10% increments.

Training deep learning-based detectors requires large, diverse, and well-annotated datasets, and assembling such a dataset for landmine detection is exceptionally difficult [20]. Access to active minefields is restricted for safety reasons, controlled surveys are expensive, and legal constraints on the handling of military ordnance further complicate systematic collection across varied soil and weather conditions [6]. Existing datasets consequently remain small and insufficiently diverse to train detection models reliably [15], [21]. In response, synthetic data generation offers a plausible remedy by rendering realistic mine imagery, allowing researchers to obtain large, labeled datasets at far lower cost and without safety risk. Models trained exclusively on synthetic data nevertheless tend to underperform because of the domain gap, since synthetic images rarely match real ones in lighting, texture, background clutter, and shadow [22]. Consequently, a common mitigation is to mix real and synthetic imagery during training, so that genuine images keep the model anchored to reality while synthetic images augment the dataset [22].

Despite deep learning's clear potential for automated demining, the field remains constrained by the scarcity and limited diversity of real-world imagery. Synthetic data generation offers a path forward, but to be effective, it must be combined with some real-world data [23]. The extent to which mixing real and synthetic samples can bridge the domain gap, and the optimal mixing ratio, remain gaps in the literature. This study directly addresses this gap in the context of multispectral landmine detection. Specifically, we investigate how varying the proportion of synthetic training imagery, while holding total training set size constant, affects detection performance on real-world data. We then evaluate this across two architecturally distinct computer vision models. These models are YOLOv11, which is a single-stage convolutional detector, and RF-DETR, a transformer-based end-to-end detector. We selected these models to determine whether they respond differently to the same real-synthetic blend. We further examine how the value of synthetic imagery varies with the conditions under which the real imagery was acquired, since the testing dataset encodes sensor altitude, time-of-day, season, and thermal fusion levels. This research will attempt to answer the following three research questions:

1) What synthetic-to-real ratio in a fixed-size training set maximizes detection performance on real-world multispectral data?

2) How do YOLOv11 and RF-DETR perform when trained on synthetic data ratios compared to real-world training data?

3) Which survey conditions (altitude, time of day, season, and thermal fusion share) benefit most from synthetic imagery, and does that pattern change when synthetic images replace real-world ones?

## II. LITERATURE REVIEW

### A. UAS-BASED LANDMINE AND UXO DETECTION

Some of the earliest deep-learning work on UAS landmine imagery trained Faster R-CNN on multispectral and thermal frames of scattered PFM-1 mines, establishing an early benchmark for the task [24]. The predecessor work behind this study follows in that line. A first comparison trained YOLOv7 on pure RGB, pure long-wave infrared (LWIR), and 50/50 fused inputs across eight altitudes and five times of day, finding that fusion edged out RGB overall and delivered its largest gains at night [18]. A broader follow-on extended the comparison to YOLOv8, YOLOv10, YOLOv11, RF-DETR, Faster R-CNN, and RetinaNet on real inert anti-tank and anti-personnel mines. YOLOv11 led the fusion-ratio sweep at 86.8% mAP at IoU 0.5, with the best operating point at 10 to 30% thermal fusion and altitudes of 5 to 10 meters [13]. The dataset behind that work, the Adaptive Multispectral Landmine Identification Dataset (AMLID), provides the real imagery used here: 12,078 labeled images spanning 21 landmine types, captured in RGB and LWIR at 11 fusion levels, four altitudes, two seasons, and three times of day [20]. Sensor fusion on demining drones predates pixel-level blending: earlier work combined several optical sensors at the decision level, setting survey altitude so that each sensor saw enough pixels on a mine to make its own detection and then merging the separate detections, without ever blending the imagery itself [25]. Labeling costs have been tackled separately by transferring Segment Anything Model masks from RGB onto co-registered multispectral frames, cutting labeling time by 87.8% at a cost of 6.7% mAP [19]. Across the wider literature, a meta-review of roughly 400 papers found RGB-LWIR to be the most common sensor pairing, at about 39%, with UAS multispectral YOLO work roughly doubling since 2020 [10]. Fusion selection has even reached deployment: one system trained 33 separate YOLOv11n models, one per fusion ratio per illumination band, and used a lux sensor to switch models in real time as illumination changed [16].

Thermal sensing has also been studied, independent of fusion ratio. Time series of thermal images outperform single frames because the temperature difference between soil and a buried mine varies predictably across the day, which informs how this study calibrates rendering parameters to time-of-day [26]. Broader reviews situate optical and thermal sensing alongside GPR, magnetometry, and neutron backscatter, and a 2024 review extends the picture to the full range of demining practice, including mechanical clearance, manual prodding, and trained animals [3],[27],[29],[30]. More recent work integrates radar and optical sensors on cooperating ground robots, and a recent aeromagnetic system reports better accuracy and greater range on buried ordnance [31],[32]. Within the optical line, one study replaced the fuse-then-detect pipeline with a joint network that learns fusion and detection together, and an RGB-only real-time baseline provides a reference for isolating thermal fusion's contribution [33],[15]. The operational need remains immediate: clearance is still underway in active conflict zones; UAV, remote sensing, and GIS practice in Ukrainian demining has been surveyed directly, and a ground-robot multi-sensor demining dataset shows how sensor payloads and deployment scope differ between ground and aerial platforms [34],[35].

## B. MULTISPECTRAL FUSION BENCHMARKS AND ARCHITECTURES

Two datasets set the evaluation conventions for RGB-thermal fusion outside the landmine domain. The KAIST multispectral pedestrian benchmark, with 95,000 aligned color-thermal pairs, set the co-registration and fusion-evaluation protocol still used a decade later [36]. LLVIP added more than 30,000 aligned pairs shot in low light and at night, and it found again that thermal fusion pays off most after dark [37]. Both were shot from the ground. DroneVehicle is the aerial equivalent, with 28,439 RGB-infrared pairs captured from a drone between day and night. The detector released with it weights each modality by a cross-modal intersection-over-union and an illumination estimate, rather than using a fixed ratio [38]. The setting is similar to this one, and the model's weight for the thermal channel changes with lighting. A re-examination of RGB-thermal fusion found that the thermal channel does not always help, and that its usefulness depends on the scene [39].

Fusion architectures have moved from channel concatenation to attention- and state-space designs. The Cross-Modality Fusion Transformer replaces concatenation with cross-attention between modalities, and it remains a standard baseline [40]. DAMSDet fuses RGB and thermal features adaptively per query instead of at a fixed ratio [41]. C2Former inserts a calibration step before cross-modal fusion for sensors that are never perfectly pixel-aligned [42]. ICAFusion applies iterative multi-round cross-attention in place of a single fusion pass [43]. Convolutional and transformer fusion have also been compared head-to-head on RGB-infrared remote sensing imagery [44]. More recent work drops both the two-modality and co-registration assumptions and learns to re-align features before fusing them. MiPa avoids a fixed ratio altogether by training one detector on randomly mixed RGB and thermal patches, so the model is modality-agnostic at inference [45]. Further refinement confines cross-attention to a local pixel neighborhood, which helps efficiency and small-object recall [46]. State-space models are the newest approach here, and a Mamba-style state-space model in place of attention-based fusion matches transformer accuracy at lower cost [17].

Vision-language models enter this study as an auto-labeling mechanism, not as a detector, but the lineage is worth setting out. CLIP put paired image and text encoders in a shared embedding space and underpins both open-vocabulary detectors and VLM-based realism metrics [47]. LoRA made adapting such models tractable without full fine-tuning [48]. Grounding DINO, YOLO-World, and YOLOE carry open-vocabulary detection into real-time operation [49], [50], [51]. One caution applies to all of them here. A 2025 review of VLM-based detection on drone imagery notes that the aerial viewpoint badly distorts the ground-level visual statistics CLIP was trained on, so web-trained models need substantial domain adaptation before they work on top-down UAS views [52]. Remote-sensing-specific models address this. GeoChat is a conversational assistant tuned to remote-sensing imagery [53]. EarthGPT handles optical, SAR, and infrared inputs in one model [54]. GeoGround extends the approach to pixel-level visual grounding [55].

## C. DETECTION ARCHITECTURE FOUNDATIONS

The detectors compared in this study descend from a small number of architectural contributions. Faster R-CNN introduced the region proposal network and is still the two-stage baseline in landmine comparisons [56]. RetinaNet's focal loss addressed the severe foreground-background imbalance in dense detectors, which matters here because mine pixels are rare against background terrain in an aerial frame [57]. The Vision Transformer showed that a convolution-free transformer can match or beat convolutional networks given enough data, and it underpins essentially every transformer detector above [58]. Deformable DETR fixed the slow convergence and weak small-object performance of the original DETR by sampling a sparse set of key locations, and it is the direct predecessor of both RF-DETR and RT-DETR [59]. RT-DETR was the first real-time end-to-end transformer detector to beat YOLO on speed and accuracy, which it did by removing non-maximum suppression [60]. YOLOv11 adds C3k2 blocks, spatial pyramid pooling fast, and C2PSA attention over earlier releases, and it is the architecture used most often in the landmine studies above [61]. SAM 2 extends promptable segmentation to video through a streaming memory mechanism, which opens a route to auto-labeling flythrough sequences instead of isolated frames [62].

## D. SYNTHETIC DATA GENERATION AND DOMAIN RANDOMIZATION

Domain randomization showed that detectors can transfer from deliberately unrealistic, randomized simulation to the real world. That is one end of the realism-versus-randomization spectrum this study traverses [63]. At the other end, BlenderProc2 is the most widely used procedural pipeline for photorealistic renders with automatic COCO- and BOP-format labels [64]. Synthetica is a different engine choice, with randomization at scale through Isaac Sim and Omniverse instead of Blender [65].

Generative methods supplement direct rendering. CycleGAN translates between domains without matched pairs, including visible-to-infrared style [66]. InfraGAN synthesizes a plausible LWIR counterpart to a visible image and beats Pix2Pix and ThermalGAN on structural similarity, making it useful wherever an LWIR channel must be produced from a rendered RGB scene rather than simulated from thermal physics [67]. Where physical fidelity is needed instead, a finite-volume heat-transfer simulation that accounts for radiation, convection, and conduction yields physically grounded surface temperatures for infrared rendering [68]. Diffusion models have since been applied directly to detector training data. AeroGen conditions generation on object layout and improves mAP by 2.4 to 4.3% across three aerial benchmarks, with the largest gains on rare classes [69]. InstaGen produces images and bounding boxes together and gains 4.5 points of open-vocabulary precision, with the strongest effect when real data is scarce [70]. Controllable diffusion with explicit position, size, and category conditioning gives tighter control over annotation accuracy than letting the model place objects on its own [71].

## E. MIXING RATIOS, REALISM, AND TRAINING UTILITY

This study addresses a question with an established shape in the general detection literature, as researchers have systematically varied the proportion of real to synthetic images. One such study found a band between 5 and 20% real data that matched training on the full real dataset, and it recovered the diminishing-returns curve any mixing-ratio study expected [72]. A companion line of work asked which generation factors matter. It isolated texture randomization, lighting, and camera pose variation, and weighed synthetic data quality against quantity [73]. A third argued that what closes the simulation-to-real gap for detectors is a rigorous training methodology, not more realistic imagery, and warned against over-investing in generator realism at the expense of training procedure [23].

In one study Fréchet Inception Distance compared feature-space statistics of real and generated images and is the near-universal reference metric [74]. Kernel Inception Distance is unbiased at small sample sizes and assumes no Gaussian feature distribution, which helps when scoring small batches per rendering condition instead of tens of thousands of images [75]. CLIPScore shows that CLIP embeddings can assess image-text correspondence without human captions, and that they track human judgment more closely than earlier metrics [76]. There is a caveat. Visually striking text-to-image output often makes poor detector training data, because visual realism and training utility are separate properties, so a strong FID or CLIPScore does not guarantee a better detector [77]. Validation therefore has to reach past distributional similarity. Grad-CAM is the base mechanism for asking which pixels drove a decision [78]. Extended to YOLO-family detectors, it yields per-detection heatmaps, which can be used to check whether a synthetically trained model attends to the same features as a real-data model.

## F. SMALL-OBJECT, CAMOUFLAGED, AND SCARCE-DATA DETECTION

Landmine detection compounds two difficulties: targets are small in the aerial frame, and labeled real data is scarce. Surveys of tiny object detection combined with domain adaptation treat both at once and name small target extent as a principal failure mode for standard architectures [79]. A later and much larger study built benchmarks specifically for small objects and ran a wide set of detectors against them. Naturally, accuracy falls as the target gets smaller, and it falls for every architecture tested [80]. Because a mine covered by soil or screened by vegetation is a camouflaged object, the camouflaged object detection literature provides directly applicable context, including edge-aware decoders and texture-difference modules [81], [82]. Few-shot detection addresses the data-scarcity half of the problem across standard, generalized, incremental, open-set, and domain-adaptive regimes [83]. The aerial case has been treated directly through cross-scale query-support alignment for small objects [84]. Text-guided few-shot aerial detection extends the same idea with language prompts [85]. On-board compute limits have driven knowledge-distilled infrared detectors sized for edge inference [86]. DOTA and VisDrone are the general aerial generalization tests used throughout the fusion and small object work above [87], [88].

Prior work has evaluated multispectral fusion ratios and detector architectures on real landmine imagery. In parallel, synthetic image generation, whether physics-based, adversarial, or diffusion-based, has matured for general aerial and infrared detection. What remains unaddressed is their intersection. Few studies pair tunable multispectral RGB-LWIR fusion with synthetic imagery in a landmine-specific setting. The realism metrics such work would rely on, FID, KID, and CLIPScore, have not been validated against downstream detection performance in this domain, and recent evidence that visual realism and training utility diverge makes that gap sharper. Three things compound the difficulty: aerial-viewpoint distortion of web-trained features, the absence of real labeled data for rare mine types, and the small, camouflaged nature of the targets. Closing the gap between synthetic training performance and real-world detection will take advances in training methods and validation, not in generator realism alone.

# III. METHODOLOGY

The study proceeds in the three stages shown in Figure 1. Stage A generates training imagery. Real AMLID frames follow one path, and three synthetic paths derive from them: Vision Language Model (VLM) assisted Blender renders, a Generative Adversarial Network (GAN) applied to the real frames, and the same GAN applied to the renders. Every source is composited into RGB-LWIR pairs at 10% fusion increments and then drawn into size-matched training sets running from 100% real to 100% synthetic. Stage B trains the detectors: sixteen training sets per architecture, comprising a real-only baseline, nine real-synthetic mixes, three pure-synthetic sets, a four-way split, a three-way synthetic split, and one combined real-synthetic pool, each used to train YOLOv11 and RF-DETR for 32 models in total. Stage C evaluates them by scoring all 32 models on the same 1,320 real held-out images, then recording results by detector, season, illumination, sensor fusion level, mine type, and flight altitudes of 5, 10, 15, and 20m. The subsections below follow that order. Sections III-A through III-D describe the four image sources and the fusion procedure of Figure 1A. Sections III-E and III-F describe the training set composition and detector configuration of Figure 1B, while Sections III-G through III-J describe the test set, metrics, and analysis of Figure 1C.

## A. SYNTHETIC SCENE GENERATION

The first synthetic pathway in Figure 1A begins with the virtual environment from which the RGB imagery was sourced, built around external models of 21 landmines: 12 anti-tank and 9 anti-personnel. Materials matched each device's real-world construction. The mines were placed on a flat, textured ground plane with density-painted grass and dirt materials, populated with procedural grass particles, and bare dirt patches distributed in a pattern matching the AMLID real-image dataset environment. A single directional lamp stood in for the sun, supplemented by a warm fill light, with solar elevation and azimuth set per time of day by the image generation script. Scene authoring was carried out with the assistance of a Claude Opus 5 vision-language model implemented through a Model Context Protocol (MCP) server. The model helped construct the scene and did not generate or modify any annotation.

Each mine was assembled from discrete component meshes instead of as a single body. The TM-62 Russian anti-tank mine, for example, was built from a body, seat ring, pressure cap, washers, a brass fuse plug, and a striker, each a separate external mesh carrying a painted-metal, bare-metal, or plastic material assignment, grouped together to form the final model. Anti-tank and anti-personnel mines were then scattered at random across the environment, all placed above ground and spaced so as not to sit too close together.

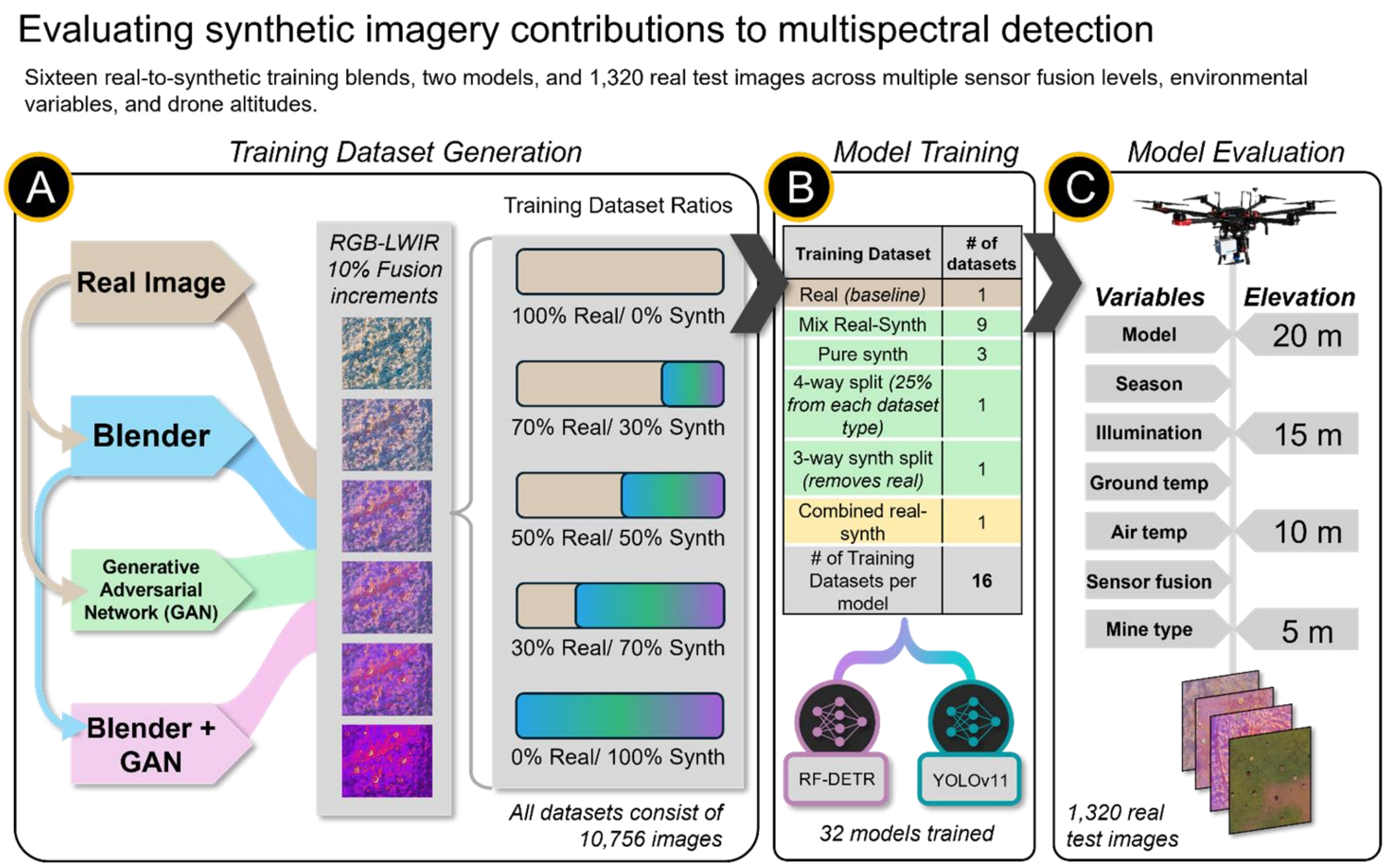

**FIGURE 1.** Study design for evaluating synthetic imagery in multispectral landmine detection. (A) Four image sources composited into RGB-LWIR pairs at 10% fusion increments and drawn into size-matched training sets from 100% real to 100% synthetic. (B) Sixteen training sets per detector, each used to train YOLOv11 and RF-DETR for 32 models. (C) All models scored on the same 1,320 real held-out images, stratified by detector, environmental condition, fusion level, mine type, and altitude (5–20 m).

## B. THERMAL SIMULATION

Every path in Figure 1A terminates in a matched RGB-LWIR pair, and the rendered scenes require a thermal channel that no render produces natively. We therefore generated synthetic LWIR imagery by layering a reduced-order thermal appearance model over the Blender renders. We converted surface geometry, per-material thermal properties, and cast shadows to apparent temperature analytically as a post-process instead of through a coupled solver, then rasterized the resulting field into a false-colored LWIR frame for downstream use. Rasters were rendered at 3× super-sampled resolutions for each camera pose and then downsampled.

Two properties of this approach matter for Section V-G. The model does not reproduce sensor-level thermal noise, and it assigns each mine one of only three blanket materials: generic painted-metal, bare-metal, or generic plastic, each carrying a uniform set of thermal constants. It therefore falls short of a physically based, finite-volume heat-transfer simulation that provides full radiometric fidelity for infrared rendering.

## C. MULTISPECTRAL FUSION AND IMAGE CAPTURE

This step produces the fusion stack at the center of Figure 1A, showing the same scene at successive levels of thermal contribution. Once the thermal simulation was finalized, an output script exported matched pairs of visible-band and synthetic-LWIR frames at 5, 10, 15, and 20 meters. The two registered bands are composed using alpha blending. This blend defines the imagery used throughout the study, and Section V-F draws on the fact that it composites two bands rather than modeling a single fused detector when interpreting fusion behavior.

## D. GAN-BASED DOMAIN REFINEMENT

The remaining two pathways in Figure 1A, GAN and GAN + VLM, share a single refinement network. To reduce the visual gap between our Blender renders and real aerial thermal imagery, we trained a CycleGAN to edit rendered frames so that they more closely resemble the real AMLID data. We adopted an unpaired method because no pixel-matched real photograph exists for any rendered scene. The cycle-consistency loss also preserves each image's content, so a mine in a refined frame remains where it was in the original render, and the original Pascal VOC annotation remains correct. Labels are therefore

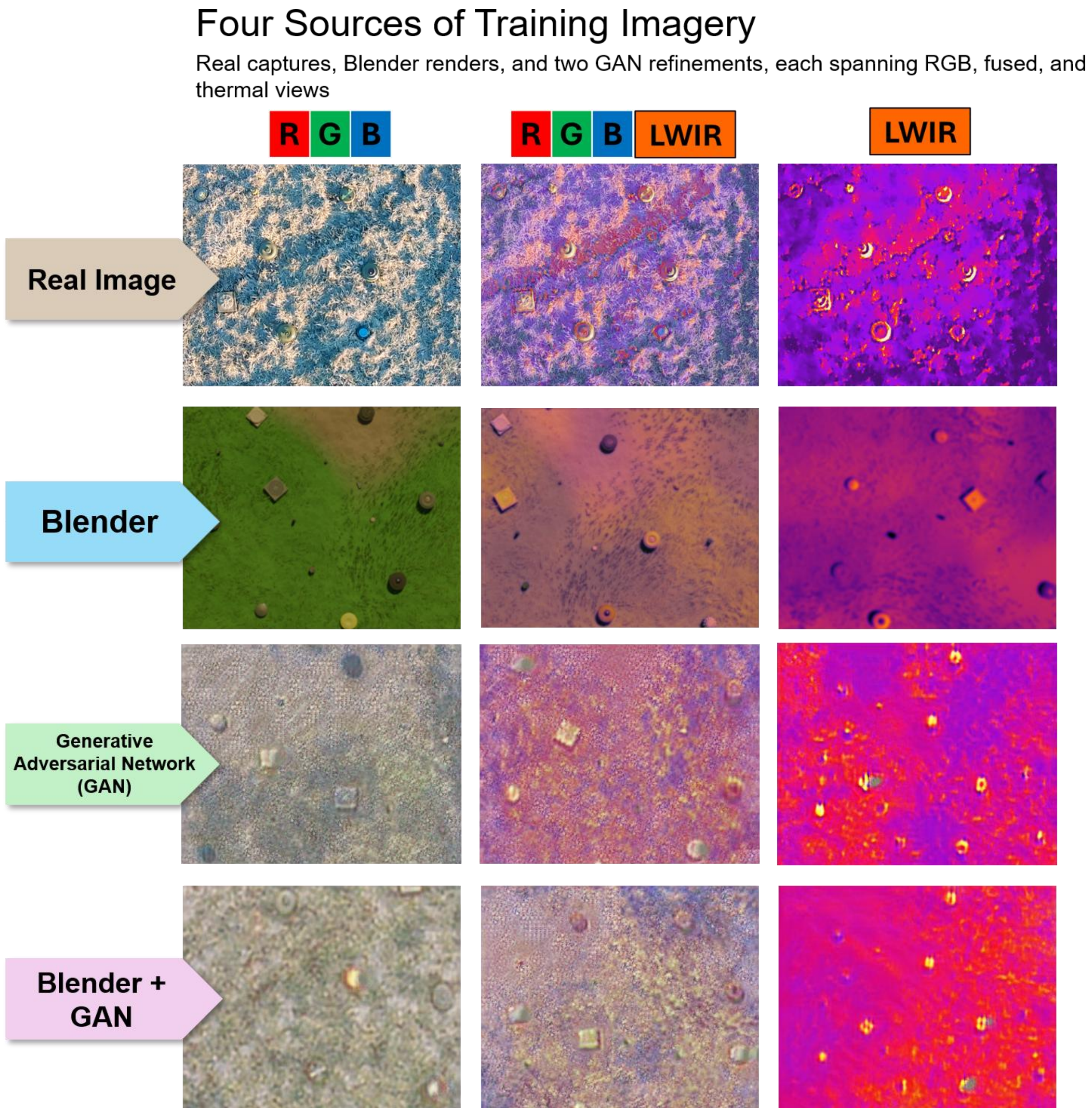

**FIGURE 2.** **The four image sources used to build the training sets, shown across the three spectral products. All images are from the January morning condition. Columns show the RGB band, the alpha-blended RGB-LWIR fusion, and the LWIR band. Rows show a real AMLID capture, a Blender render, a GAN-refined real frame (GAN-Real), and a GAN-refined render (GAN-Blender). Each source supplies matched visible and thermal channels, so training mixes can be composed at any fusion level. Scenes differ between rows; the panels illustrate the appearance of each source rather than a registered comparison of the same scene.**

copied from the rendered images to the refined images unchanged.

The source domain comprises 978 rendered scenes, each stored as a separate visible-band image and false-colored LWIR image. The target domain comprises 7,517 real AMLID frames. For each synthetic frame, we draw a real reference from the same season, time of day, and fusion level. Importantly, we excluded the held-out elevation test images and their refined copies from the target domain so no test data could influence training.

Both generators and discriminators receive a four-value condition vector describing fusion level, season, time of day, and altitude. Each value is expanded into a constant channel and stacked onto the image channels. Conditioning the discriminator proved as consequential as conditioning the generator: when only the generator received the condition, the network settled on a single average appearance and lost the color differences between conditions. Conditioning also lets the generator be steered, so it can produce imagery for any combination of season, time, altitude, and fusion level.

The generators are ResNet-style networks (48 base filters, six residual blocks, 256 × 256 input) that upsample by resize-plus-convolution instead of transposed convolution. The discriminators are 70 × 70 PatchGAN critics [89]. The loss combines a least-squares adversarial term [90] with cycle-consistency ($\lambda$ = 10), identity ($\lambda$ = 5), and feature-matching ($\lambda$ = 2) terms [91]. Because the real dataset is small, discriminator inputs pass through ADA-style differentiable augmentation with an adaptive probability [92]. We restricted this to geometric transforms; when we included brightness and contrast jitter, the generator produced washed-out images that nonetheless satisfied the discriminator.

We trained with batch size 1 using Adam (learning rate $2 \times 10^{-4}$, $\beta_1 = 0.5$) for 28,800 iterations, decaying the learning rate linearly from the halfway point, on an NVIDIA Jetson AGX Orin, requiring approximately eight hours per run. We selected the best checkpoint using a combined Earth Mover's distance between the refined and real image histograms, measuring grayscale intensity, per-channel color, and Sobel gradient magnitude on a fixed set of sample images.

We then ran the refiner over every rendered frame (10,758 images) and, as a control, over every real frame (12,078 images). Both sets replicate the folder structure and per-folder image counts of the real dataset. For the few conditions in which the real reference uses a different imaging product, we applied a per-condition intensity calibration. Overall, refinement lowered the mean Earth Mover's distance against real imagery from 31.95 to 5.74.

### E. TRAINING SET COMPOSITION

The ratio bars in Figure 1A and the dataset tally in Figure 1B describe the same sixteen sets from two directions: the bars show how real and synthetic imagery trade off within a fixed set size, and the tally shows how the sixteen sets are divided by type. We assembled these sixteen training sets to test how synthetic imagery affects landmine detection on real data. Each set draws from four image sources: real AMLID frames, raw Blender renders, GAN-refined Blender renders, and GAN-refined real frames. The sets are organized into four groups, each addressing a different question.

Group 1 (mixes 1 to 4) trains on a single source at a time (real only, Blender only, GAN-Blender only, and GAN-Real only) to establish a baseline for each source in isolation. Group 2 (mixes 5 to 13) measures response curves: for each of the three synthetic sources, we build three mixes holding 70, 50, or 30% real imagery, so that performance can be tracked as real data is progressively replaced. Group 3 (mixes 14 to 15) tests whether the sources complement one another; mix 14 splits the data evenly across all four sources at 25% each, while mix 15 removes real imagery entirely and splits evenly across the three synthetic sources, establishing a floor for what synthetic data alone can achieve. Group 4 (mix 16) pools every available image from all four sources to test whether additional data improves performance regardless of composition.

Mixes 1 through 15 each contain the same number of images (10,756), with composition varying as shown in Table 1. Holding size constant makes composition the only variable, so any performance difference can be attributed to the synthetic-to-real ratio and to synthetic type, not to the volume of data the model saw. A 70/30 mix, for example, draws 7,529 real and 3,227 synthetic images. Mix 16 is the exception: it pools all imagery from every source and is therefore roughly four times larger than the others. Because it is not size-matched, it answers a different question and is not directly comparable to the rest.

Images are drawn at random with a fixed seed per mix so that the sets can be rebuilt exactly, and each is split 80/20 into training and validation partitions. We train all models on two mine classes: anti-personnel (AP) and anti-tank (AT). The real annotations subdivide these by casing material, metal or plastic, but the synthetic annotations do not carry that distinction, so we collapsed the real labels to the two shared classes to keep every model comparable.

### F. DETECTOR TRAINING CONFIGURATION

We used each mix to train the two models shown at the foot of Figure 1B, YOLOv11 and RF-DETR, for a total of 32 models. YOLOv11 starts from the pretrained yolo11n checkpoint and trains at 640 × 640 with batch size 16, initial and final learning rates of 0.01, and the default Ultralytics augmentation pipeline. RF-DETR, in turn, starts from the pretrained RFDETRBase checkpoint and trains with batch size 4 and four gradient-accumulation steps, giving an effective batch of 16, a learning rate of $1 \times 10^{-4}$, and early stopping with a patience of 10. Both models train for 50 epochs. Training ran on NVIDIA A100 GPUs through Google Colab, and for each model we retained the best checkpoint by validation performance.

### G. TEST SET AND LEAKAGE CONTROL

Evaluation, Figure 1C, is common to all 32 models. All models were evaluated against a single held-out test set consisting of the 1,320 real images in the AMLID elevation test folder. These images appear in no training or validation split, and the same test set was used for every model, so differences in reported performance are attributable to training composition, not evaluation difficulty. Crucially, no synthetic image appears in the test set at any point, which means every reported figure measures performance on real sensor imagery, the quantity of operational interest.

We also controlled for a second, less obvious form of leakage. Because the GAN-Real pool consists of real AMLID frames passed through the refinement network, the refined counterparts of the test images depict the same scenes with the same mines in the same positions, differing only in style.

Training on those counterparts would expose a model to test content in a form that a naive file-level check would not detect. The refined counterparts of all 1,320 test images were therefore removed from the GAN-Real pool before any mix was assembled. Similarly, we excluded the elevation test images from the target domain used to train the refinement network, so no test content influenced the generator.

### H. MODELS EVALUATED AND METRICS

All 32 models were run over the complete test set. YOLOv11 was selected because it produced the strongest results among the YOLO variants compared in the predecessor study on this dataset, as reviewed in Section II-A. We selected RF-DETR as an architecturally distinct comparison because it is a transformer-based end-to-end detector that dispenses with non-maximum suppression, and because transformer and two-stage architectures behaved differently from single-stage detectors under altitude and fusion variation, as reviewed in Section II-C.

**TABLE 1**
COMPOSITION OF THE SIXTEEN TRAINING SETS

| # | Training set | Real | Blender | GAN-Blender | GAN-Real | Total |
|---|---|---|---|---|---|---|
| 1 | Real only | 10,756 | 0 | 0 | 0 | 10,756 |
| 2 | 100% Blender | 0 | 10,756 | 0 | 0 | 10,756 |
| 3 | 100% GAN-Blender | 0 | 0 | 10,756 | 0 | 10,756 |
| 4 | 100% GAN-Real | 0 | 0 | 0 | 10,756 | 10,756 |
| 5 | 30% Blender | 7,529 | 3,227 | 0 | 0 | 10,756 |
| 6 | 50% Blender | 5,378 | 5,378 | 0 | 0 | 10,756 |
| 7 | 70% Blender | 3,227 | 7,529 | 0 | 0 | 10,756 |
| 8 | 30% GAN-Blender | 7,529 | 0 | 3,227 | 0 | 10,756 |
| 9 | 50% GAN-Blender | 5,378 | 0 | 5,378 | 0 | 10,756 |
| 10 | 70% GAN-Blender | 3,227 | 0 | 7,529 | 0 | 10,756 |
| 11 | 30% GAN-Real | 7,529 | 0 | 0 | 3,227 | 10,756 |
| 12 | 50% GAN-Real | 5,378 | 0 | 0 | 5,378 | 10,756 |
| 13 | 70% GAN-Real | 3,227 | 0 | 0 | 7,529 | 10,756 |
| 14 | Equal synthetic | 2,689 | 2,689 | 2,689 | 2,689 | 10,756 |
| 15 | All synthetic | 0 | 3,586 | 3,585 | 3,585 | 10,756 |
| 16 | Real + all synthetic | all | all | all | all | ≈43,000 |

**TABLE 1: Mixes 1–15 are size-matched at 10,756 images; each label names the synthetic share. Mix 16 pools every available image from all four sources.**

Evaluating both architectures on identical mixes separates effects attributable to training data from effects attributable to architecture.

Mean average precision at an intersection-over-union threshold of 0.5 is the primary metric, chosen for direct comparability with the predecessor study and with the wider landmine detection literature, which reports mAP@0.5 almost universally. We also report mean average precision averaged across thresholds from 0.5 to 0.95, since the two can diverge when a detector reliably registers a target but bounds it imprecisely, and coarse localization carries different operational consequences than outright failure to detect.

Precision, recall, and F1 are reported at a confidence threshold of 0.25, which the team adopted as a common operating point to keep results from different members comparable. Recall carries particular weight in interpretation because the two error types are not symmetric in a demining context: a false positive cost verification effort, whereas a false negative leaves live ordnance in ground that has been surveyed and may be declared clear.

The two metric families are computed from a single inference pass but at different thresholds, and the distinction matters. Mean average precision is computed over the full detection set, retaining low-confidence detections so the precision-recall curve is complete, following the standard COCO protocol. By contrast, computing mAP after discarding detections below 0.25 would truncate the tail of that curve and depress the resulting figure, making the values reported here incomparable with both the predecessor study and the published literature. We then compute precision, recall, and F1 from the same detection set filtered to the 0.25 operating point. Reporting both preserves comparability with prior work while retaining the operating point the team standardized on.

We report average precision separately for the anti-personnel and anti-tank classes, not only in aggregate. Because the class distribution is heavily weighted toward anti-tank mines and the two classes differ substantially in pixel extent, an aggregate figure conceals the disparity between them, and that disparity is among the more consequential findings of this study.

## I. STRATIFIED EVALUATION

Beyond aggregate performance, we computed results within strata along four dimensions listed in the variables and elevation columns of Figure 1C: sensor altitude (5, 10, 15, and 20 m steps), time of day, RGB-LWIR fusion level, and season. We evaluated each stratum independently, filtering detections to images belonging to that stratum before computing metrics, and we never pooled fusion levels within a single evaluation bucket. Aggregate reporting alone, however, would average across conditions in which the two sensor modalities contribute very differently, obscuring precisely the interactions this study is designed to measure. Detection accuracy on this dataset varies by more than 40% across altitude alone, as reviewed in Section II-A, which establishes that these strata are not incidental.

We wrote results to a common tabular schema so outputs from both teams concatenate without reconciliation. Each row records the model identifier, architecture, stratum, level within that stratum, image and ground-truth box counts for the bucket, mAP@0.5, mAP@0.5:0.95, and precision, recall, and F1 at the 0.25 operating point. Because every mix is evaluated by both architectures and the mixes within a group are directly comparable, the spread across models within a stratum provides the standard error bars reported in the results figures.

## J. ANALYSIS PROCEDURE

We plotted performance for each architecture against the real-data fraction of the training set across the size-matched mixes, producing a response curve for each synthetic source independently. Each mix was additionally normalized against the real-only baseline of mix 1, so that a relative value of 1.0 denotes parity with training on real data alone. This normalization lets us compare the two architectures directly, even when their absolute mAP differs. We then compared the three synthetic sources at matched real-data fractions, which isolates the generation method's contribution from the mixing ratio, and compared mix 16 against mix 1 to separate the effect of data volume from the effect of data composition.

A third analysis addresses the survey variables directly. For each stratum already encoded in the test set, namely sensor altitude, time of day, season, and thermal fusion share, the difference in mAP@0.5 between a given mix and that architecture's real-only baseline was computed cell by cell and then averaged within the stratum, yielding a per-condition delta instead of an aggregate one. Two contrasts were run in this form. We measured substitution using the size-matched mixes, which hold total set size constant so that variation in the delta across strata reflects the acquisition condition rather than the amount of data. Augmentation, by contrast, was measured using mix 16, which adds all synthetic imagery to the full real set. Because the two contrasts are computed against the same baseline in the same strata, they can be compared directly, allowing the substitution and augmentation responses to be read against one another rather than in isolation.

In sum, the design yields 32 trained detectors, two architectures across sixteen training sets, all scored on the same 1,320 held-out real images spanning 264 acquisition conditions. The results follow the order of the research questions: substitution ratio first, architecture comparison second, and survey conditions third.

## IV. RESULTS

### A. SUBSTITUTION RATIO IN A FIXED-SIZE TRAINING SET (RQ1)

Figure 2 reports the study as three rows of two panels, one row per research question. The first asks which synthetic-to-real ratio in a fixed-size training set maximizes detection performance on real-world multispectral data, and how much real imagery can be displaced before accuracy declines. The top row addresses it, with total set size held constant at 10,756 images so that composition is the only variable. 2A plots accuracy against the synthetic share of the training set for all three synthetic sources. No substituted set exceeds either real-only baseline of 73.1 mAP@0.5 for RF-DETR and 71.4 for YOLOv11, so the maximum falls at a synthetic share of zero. Blender renders and imagery refined from real frames nevertheless hold close to those baselines well up the ladder: RF-DETR reaches 71.4 at a 50% Blender share and 71.4 at a 30% GAN-Real share, and YOLOv11 reaches 69.9 at a 30% Blender share, recovering between 97.7 and 98.0% of the corresponding baseline. By contrast, imagery refined from renders behaves differently at every share, costing 17.0 and 19.4 points at 30% share for RF-DETR and YOLOv11, versus 2.8 and 1.4 points for the equivalent Blender substitution. At a fully synthetic composition, however, every source collapses to 36.5 and 23.0 mAP@0.5 for Blender, 54.7 and 32.8 for GAN-Real, and 23.0 and 14.1 for GAN-Blender. One point departs from the pattern and is ringed in the panel: RF-DETR at a 70% GAN-Blender share records 70.7, against 56.1 and 55.8 at the 30 and 50% shares. The inversion appears in the underlying results rather than in the analysis, and we treat this single point as a probable training artifact rather than a systematic effect.

Figure 2B holds the same five principal training sets against flight altitude, pooling both detectors. Every configuration degrades as ground sampling distance grows, but not at the same rate. Real-only imagery falls from 85.7 mAP@0.5 at 5 m to 59.7 at 20 m, and the fully augmented set tracks it almost exactly, at 84.8 and 61.6. The purely synthetic sets diverge sharply: Blender falls from 51.8 to 10.5, GAN-Real from 54.7 to 33.7, and GAN-Blender from 49.5 to 1.0. Rendered and render-refined sets therefore shed 80 and 98% of their 5 m accuracy across the altitude range, against 30% for the real-anchored sets. Imagery refined from real frames is the exception, losing only 38%: it is the only fully synthetic source that degrades gracefully with altitude, which suggests that retaining real scene content preserves resolution-dependent detail even when the sensor domain is synthesized. A single tolerance figure therefore understates the problem, because what a given synthetic share costs depends on where in the survey envelope it is spent. The middle row turns first to the detectors, and the bottom row returns to that envelope directly.

### B. DETECTOR COMPARISON ACROSS TRAINING SETS (RQ2)

The second research question asks how YOLOv11 and RF-DETR perform when trained on synthetic rather than real imagery. The middle row of Figure 2 addresses it, comparing the two detectors on every training set and then across the temporal conditions.

Figure 2C places both detectors on every training set, grouped so that the shares within each source family sit together. Results span 62.2 points, from 14.1 mAP@0.5 for YOLOv11 trained entirely on GAN-Blender imagery to 76.3 for YOLOv11 trained on the full real set augmented with all synthetic sources. The two detectors peak on different sets: RF-DETR peaks on real-only imagery at 73.1, whereas YOLOv11 peaks on the augmented set at 76.3, which is 6.9% above its own real-only baseline and 4.4% above the best RF-DETR result. That augmented set, shaded in the panel, is the only one of the sixteen on which either detector exceeds its real-only baseline. Notably, RF-DETR leads on real imagery by 1.7 points, and the separation broadly widens as the synthetic share rises, reaching 13.5 points on the fully synthetic Blender set, 18.7 on the pooled all-synthetic set, and 21.9 points on imagery refined from real frames. The ordering reverses exactly two sets, the equal-share set and the fully augmented set, with YOLOv11 ahead by 3.2 and 6.5 points, respectively. Only the latter augments: the equal-share set is size-matched substitution, so the reversal is not a property of augmentation alone.

Figure 2D resolves the five principal sets across the six season and time-of-day combinations, again pooling both detectors. Morning acquisition is strongest for the real-anchored sets in January, at 88 mAP@0.5 for real-only imagery and 85 for the augmented set, but the ordering reverses in May, where afternoon leads at 81 and 78 against 71 and 76 in the morning; across all six cells the two sets track one another closely. The purely synthetic sets are both lower and differently ordered: Blender peaks in January afternoon at 39 while GAN-Real peaks in January morning at 57, and GAN-Blender stays between 9 and 25 throughout. Season produces no consistent effect in aggregate, since January and May differ by less than a point in the real-only models, but the interaction is not uniform across sources, because Blender loses 12 to 19 points in every time-of-day cell moving from January to May, while the real-anchored sets show cell-level swings of up to 17 points in the morning and 25 in the afternoon that run in opposite directions and cancel in aggregate. In other words, temporal condition matters more for what the training imagery is made of than for which detector consumes it, which brings the analysis to the survey envelope itself.

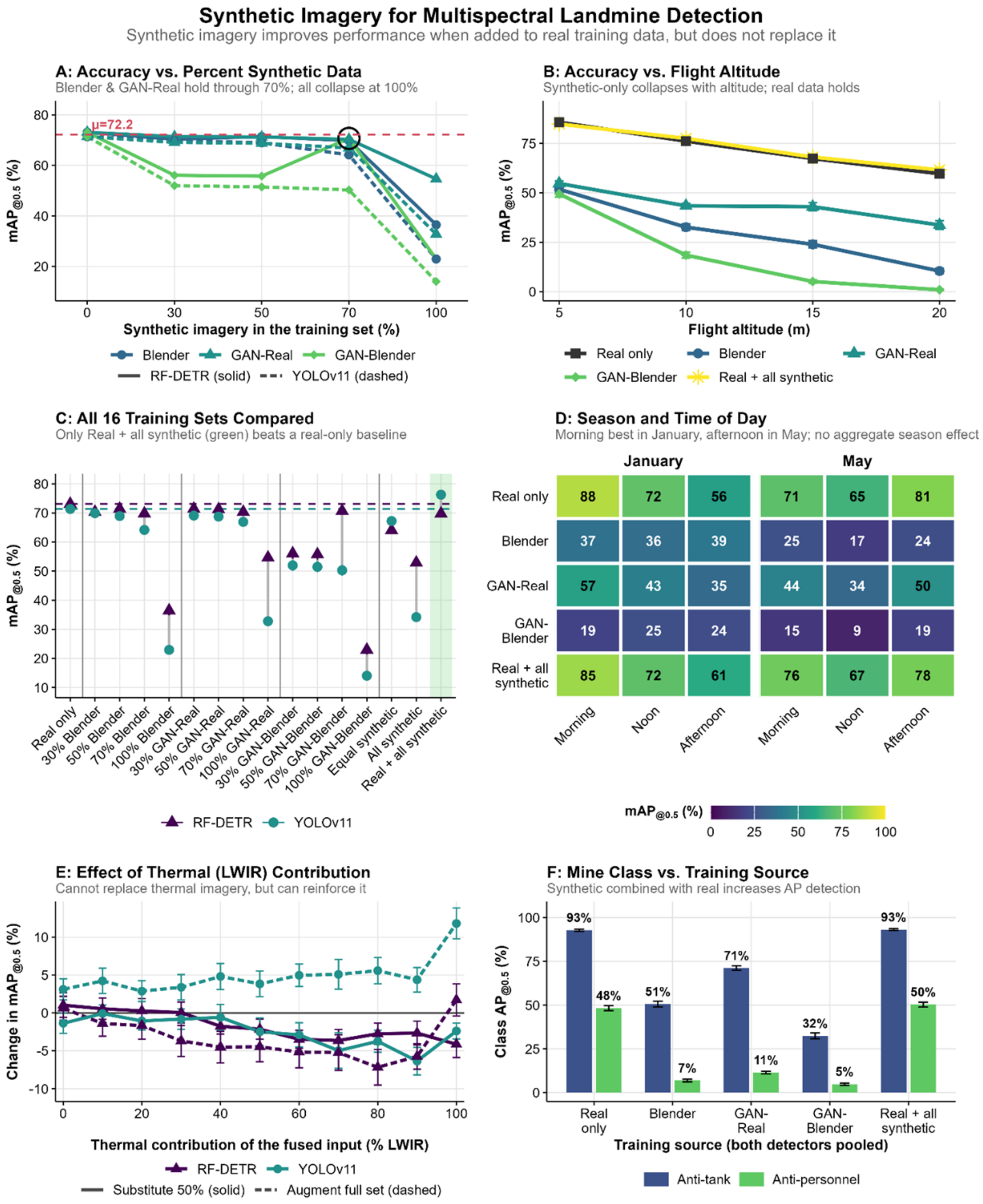

FIGURE 3. **Synthetic imagery for multispectral landmine detection across 32 models and 264 test conditions. (A) accuracy against synthetic share, by source; (B) accuracy against flight altitude; (C) all sixteen training sets, by detector; (D) accuracy by season and time of day; (E) change in accuracy against thermal contribution, substitution versus augmentation; (F) class average precision by training source.**

### C. SURVEY CONDITIONS AND CLASS PERFORMANCE (RQ3)

The third research question asks where synthetic imagery contributes within the survey envelope, and whether that pattern differs between substituting for real imagery and augmenting it. The bottom row of Figure 2 addresses this question along the two axes: the thermal contribution of the fused input and the mine class.

Figure 2E plots the change in accuracy against the thermal contribution of the fused input for a 50% substitution and for augmentation of the full real set. The two contrasts separate cleanly. Substitution is free when the input is RGB-dominant and costly when it is thermal-dominant: RF-DETR gains 1.0 and 0.6 points at 0 and 10% thermal contribution, stays level at 20 and 30%, then declines to a loss of 4.1 points at pure LWIR. YOLOv11 runs from a loss of 1.4 points at 0% to 6.3 points at 90%. Augmentation inverts that gradient. YOLOv11 improves at every thermal level and improves most at the thermal extreme, gaining 11.8 points at pure LWIR against 2.9 to 4.2 points in the RGB-dominant band, while RF-DETR is negative across the intermediate levels and turns positive only at the two ends of the range, at 0.6 points at 0% thermal and 1.7 at pure LWIR. Both detectors therefore register their most favorable augmentation response at the single condition where substitution is least tolerable. Since the LWIR channel is the simulated channel described in Section III-B, the reading is that rendered thermal imagery carries usable signal but cannot stand in for the real measurement.

Figure 2F resolves the same sources by mine class, pooling both detectors. The class disparity dominates every other effect in the study. Real-only training yields 92.8 mAP@0.5 on anti-tank mines against 48.3 on anti-personnel, and no synthetic source recovers the smaller class: Blender reaches 50.6 and 6.9, GAN-Real 71.2 and 11.4, and GAN-Blender 32.4 and 4.7. Augmentation is the only configuration that improves real imagery for either class, reaching 93.2 on anti-tank and 50.3 on anti-personnel, a gain of 0.4 and 2.0 points. The panel therefore pinpoints where synthetic imagery actually helps: it does not close the anti-personnel gap, but when added to the real training set, it is the only intervention tested here that improves the smaller class.

Taken together, the three rows give conditional answers rather than a single verdict. Synthetic imagery can replace roughly half of a real training set at a cost of under three points, but only at low altitude where RGB dominates the fusion. The two detectors diverge as the synthetic share rises, and only augmentation beats the real-only baseline. Synthetic imagery strengthens the thermal channel and the anti-personnel class, but it cannot replace real data in either role. These conditions define a practical operating envelope: at 10 m or below, with a thermal contribution of 20% or less, on a morning survey, performance averages 74.4 mAP@0.5 and 54.7 on anti-personnel, versus 57.4 and 32.8 outside it. Within that envelope, the dominant failure is missed detections rather than false alarms: recall falls 23.9 points across the altitude range, while precision falls 17.5. The discussion that follows examines why these boundaries arise and what they mean for fielding detectors trained on synthetic data.

## V. DISCUSSION

This study first assessed how detection performance on real multispectral imagery changes when synthetic imagery replaces real imagery within a fixed-size training set, across three separately generated synthetic sources. Second, this research quantified the mAP of two architecturally distinct detectors, YOLOv11 and RF-DETR, trained on those same compositions, to determine whether the two respond similarly to a given real-synthetic blend. Third, this study resolved both effects across the survey conditions encoded in the test set, separating the case in which synthetic imagery replaces real imagery from the case in which it is added to it. The discussion will now return to the three research questions identified earlier to examine key findings.

### A. SUBSTITUTION RATIO AND DIMINISHING RETURNS (RQ1)

*1) What synthetic-to-real ratio in a fixed-size training set maximizes detection performance on real-world multispectral data?*

Across the size-matched sets, no synthetic share improved on the real-only composition. Performance peaked at 100% real, at 73.1 mAP@0.5 for RF-DETR and 71.4 for YOLOv11, and held within 2.8 points of those values through a 50% synthetic share for the rendered and refined-real sources. The operative finding is therefore a tolerance band, not an optimum: up to half of a fixed-size training set can be synthetic at limited cost, while substitution beyond a 70% synthetic share is not viable.

The finding that detection performance on real imagery does not increase monotonically with the proportion of synthetic training data is consistent with the established literature on synthetic-to-real mixing. Prior work that systematically varied the ratio of real to synthetic images within a fixed-size training set modeled the relationship as a power law and reported that mixtures containing between 5 and 20% real data reduced the requirement for real imagery most effectively without degrading detection performance, cutting the need for real data by as much as 70% [72]. The finding that transfers across domains is the shape of the curve: performance holds on a broad plateau as the synthetic share rises, then declines, rather than trending steadily in either direction. The response curves here reproduce that plateau-then-decline shape, though not its extent.

The extent of that plateau, however, disagrees with the prior result, and the disagreement is worth stating outright. That prior work reports that 5 to 20% real data suffices, implying that 80 to 95% of a training set can be synthetic at no cost. The present data do not support that in this domain. The size-matched mixes sample real-data fractions at 100, 70, 50, 30, 25, and 0%, and while substitution up to a 50% synthetic share costs under 2.8 mAP points, a 70% synthetic share already costs 3.30 points for RF-DETR and 7.19 for YOLOv11, and a fully synthetic set costs 36.62 and 48.42 points. The collapse between a 30% real fraction and a 0% real fraction is evidence against an optimum lying inside the unsampled 5 to 20% region, not merely silence about it. Three differences plausibly account for the divergence. The targets here are small, low-contrast objects occupying a few pixels in an aerial frame, not vehicle-scale objects at close range. The synthetic imagery is generated at 256 by 256 pixels and upsampled, which imposes a detail ceiling discussed in Section V-E. Finally, the LWIR channel is a physical measurement in which the detection signature resides rather than a stylistic property of the image, so rendering error in that channel is not nuisance variation. The prior study also used purely RGB ground-level imagery, whereas this study uses synthetic aerial nadir imagery in both RGB and LWIR. On balance, the plateau-then-decline shape generalizes while its width does not: the tolerable synthetic fraction is domain-specific and must be established per sensor and target class. Sampling real-data fractions between 0 and 30% would locate the knee more precisely and is an inexpensive extension of the existing design.

The expectation that synthetic data delivers its largest benefit where real data is scarcest also agrees with findings from diffusion-based synthetic data work, in which generated training imagery improved detection precision most substantially in low-data regimes and produced the largest gains on rare object classes [69], [70]. The landmine domain sits squarely in that regime, since access to active minefields is restricted and controlled surveys are costly.

Separately, mix 16 provides a control that mixing studies frequently omit. Because it pools every available image from all four sources rather than substituting one source for another, it is roughly four times larger than the size-matched mixes and is deliberately not comparable to them on composition. Its value lies in separating two explanations that a size-matched design alone cannot distinguish: whether an observed gain reflects the character of the added imagery or merely the quantity of it. The result runs in both directions: the same fourfold enlargement that improves YOLOv11 by 4.92 points over its real-only baseline (76.3 against 71.4) degrades RF-DETR by 3.33 points (69.8 against 73.1), so neither volume nor composition dominates unconditionally; the value of pooling every source depends on the architecture receiving it.

## B. REALISM VERSUS TRAINING UTILITY (RQ1)

This supports the methodological argument that a rigorous training procedure, rather than incremental improvement in generator fidelity, is what closes the sim-to-real gap for detectors; prior work has cautioned specifically against over-investing in generator realism at the expense of training methodology [23].

The question of whether physical fidelity or deliberate randomization transfers better connects to the original domain randomization result, which showed that detectors trained purely on simulated images with randomized textures and lighting can transfer successfully to real environments [63]. That position was never unconditional. A systematic analysis of which generation factors actually matter found that the value of randomizing a given parameter depends on whether it carries task-relevant signal or merely nuisance variation: randomizing a nuisance variable builds invariance, while randomizing a variable that carries the detection signal destroys it [73]. This distinction is sharper in multispectral landmine detection than in RGB-only work, because the LWIR channel is not a stylistic property of the image but the physical measurement in which the target signature resides. Thermal contrast between ordnance and soil is the detection cue; it varies predictably across the diurnal cycle, and casing material systematically alters it.

Importantly, the stratified analysis in Section IV-C supplies direct evidence on this point, and it favors the signal-bearing reading of the LWIR channel. The penalty for substitution is negligible when the fused input is RGB-dominant and grows steadily with thermal share, reaching 4.1 points for RF-DETR at pure LWIR and 6.3 for YOLOv11 at a 90% thermal share, and 12.3 points for YOLOv11 once 70% of the set is substituted. A generator whose thermal band carried the same information as the real sensor would not produce that gradient. The complementary result is that augmentation behaves in the opposite direction, improving YOLOv11 by 11.8 points at pure LWIR against roughly 3 to 4 points in the RGB-dominant band, so the rendered thermal channel evidently carries some usable signal even where it cannot stand in for the real measurement. Taken together, these two findings suggest that the realistic-versus-randomized question is better posed per parameter and per acquisition condition than as a single global choice of rendering philosophy, and that for the thermal channel specifically the priority is fidelity, not variation.

## C. COMPOUNDING DOMAIN SHIFT ACROSS GENERATIVE STAGES (RQ1)

The result that imagery produced by applying the refinement network to Blender renders underperformed both raw Blender renders and refined real imagery, while those two single-stage sources performed comparably to one another, is consistent with the domain adaptation literature. Each transformation between a source domain and the target sensor domain introduces its own distribution shift, and the GAN-from-Blender pipeline is the only condition in which two such shifts accumulate. Surveys of domain adaptation combined with tiny object detection describe this compounding behavior, in which successive shifts degrade transfer more than any single shift in isolation [79]. The ordering of the three sources matches the number of generative stages separating each from the real sensor domain, a stronger agreement than consistency alone would require.

This does not contradict the established practice of refining simulator output with an adversarial network, but it clarifies the conditions under which that practice succeeds. A secondary factor specific to this implementation is that the network was trained to map an earlier mine dataset onto the AMLID domain and was subsequently applied to the updated Blender dataset. Unpaired image-to-image translation learns a mapping between two specific domains and carries no guarantee of behavior on a third, so this mismatch represents a further shift beyond the two intended ones. The residual green and teal cast observed within mine bounding boxes is consistent with that mismatch rather than with a defect in the translation objective, particularly given that the bounding boxes in the earlier dataset were substantially larger, which would have shifted the region over which the mask preservation term was principally exercised. The cost has since been greatly reduced, though not eliminated.

Relatedly, work on translating visible imagery into the infrared domain provides context for the thermal channel specifically and reports competitive structural similarity relative to earlier paired translation methods [67]. The present results do not contradict those findings, since the degradation observed here is associated with stacking generative stages rather than the quality of any single translation step.

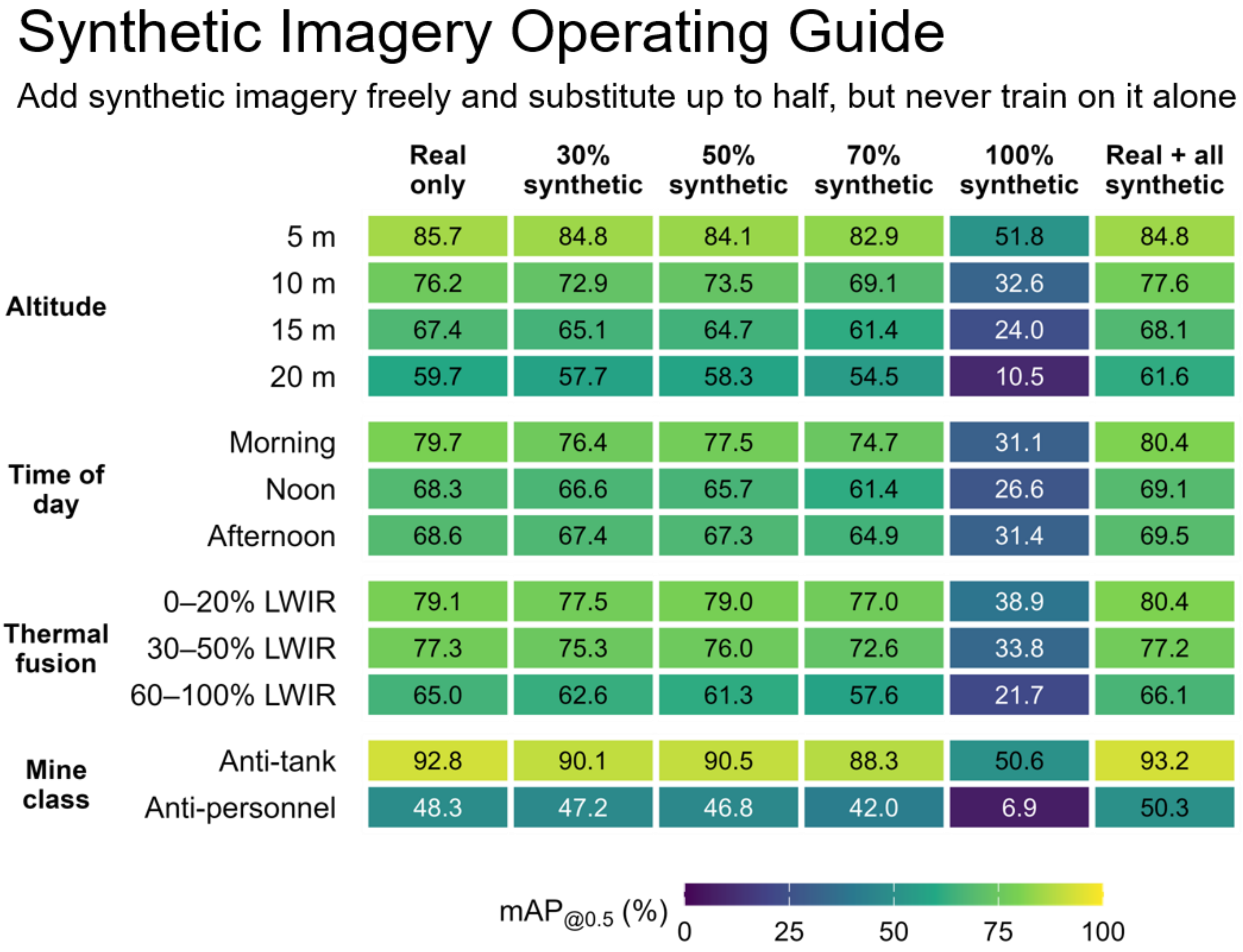

**FIGURE 4.** **Operating guide for synthetic imagery across the survey envelope. Rows are operating conditions grouped by altitude, time of day, thermal fusion band, and mine class; columns are training strategies, with substitution shares drawn from Blender renders, the strongest substitution source. Each cell reports mAP@0.5 (mine-class rows: class AP) pooled over both detectors and all remaining test conditions. Augmentation with all synthetic sources matches or exceeds the real-only baseline in every condition, substitution up to 50% costs no more than 3.7 points, and fully synthetic training collapses everywhere. GAN-Real substitution tracks the Blender results closely; GAN-Blender falls below them.**

### D. ARCHITECTURAL COMPARISON: YOLOv11 AND RF-DETR (RQ2)

*2) How do the YOLOv11 and RF-DETR detectors perform when trained on synthetic data compared to results from real training data?*

The two detectors responded differently to the same shift in training composition. RF-DETR held the higher real-only baseline, 73.1 against 71.4 mAP@0.5, and degraded more gradually under substitution, losing 3.30 points at a 70% synthetic share against 7.19 points for YOLOv11. The augmentation effect then reversed, since pooling all four sources improved YOLOv11 by 4.92 points while costing RF-DETR 3.33 points. Neither architecture is unconditionally preferable; the choice depends on whether synthetic imagery is substituted for real imagery or added to it.

The comparison between YOLOv11 and RF-DETR should be interpreted against a precedent established on this same dataset. The predecessor study [13] found RF-DETR achieving the highest overall accuracy at 69.2% mAP@0.5, followed by Faster R-CNN at 67.6%, YOLOv11 at 64.2%, and RetinaNet at 50.2% when all four were trained for 15 epochs, but also found that YOLOv11 trained 17.7 times faster than RF-DETR, at 41 minutes against 12 hours. That accuracy-efficiency tradeoff frames the present results, and the longer 50-epoch schedule used here gives both architectures more opportunity to converge than the predecessor comparison allowed.

More consequential than the ranking itself is the relative magnitude of architectural and data effects. That same study observed a 4.9% gap between the strongest transformer detector and YOLOv11 against a 56% swing driven by temporal training conditions and concluded that environmental data diversity and training strategy exert greater influence than architectural selection. Any difference observed here between the two architectures should be weighed against that precedent before being treated as decisive, and the more informative question is whether the two architectures respond differently to the same shift in training composition, not which achieves the higher absolute figure.

Divergent responses are expected. The predecessor study also found architecture-specific fusion preferences, with RF-DETR and YOLOv11 peaking on RGB-dominant input while Faster R-CNN and RetinaNet tolerated higher thermal contributions and found that the two-stage architecture degraded least with altitude. Transformer detectors attend globally instead of through a fixed convolutional receptive field, which plausibly alters sensitivity to the low-frequency style differences that distinguish rendered from real imagery. Where the two architectures rank mixes in the same order, the mixing effect can be attributed to training composition; where they diverge, the effect is architecture-dependent and should be reported as such.

### E. CLASS DISPARITY, RESOLUTION, AND SMALL-OBJECT DETECTION (RQ3)

*3) Which survey conditions (altitude, time of day, season, and thermal fusion share) benefit most from synthetic imagery, and does that pattern change when synthetic images replace real ones?*

The value of synthetic imagery was not uniform across the survey envelope, and its pattern reversed between the two cases. Under substitution the penalty was negligible where the fused input was RGB-dominant and grew steadily with thermal share, reaching 4.1 points for RF-DETR at the thermal extreme and 6.3 for YOLOv11 at a 90% thermal share. Under augmentation the same axis ran the other way, improving YOLOv11 by 11.8 points at pure LWIR against roughly 3 to 4 points in the RGB-dominant band. The disparity between anti-tank and anti-personnel targets, and the resolution at which the synthetic imagery was generated, condition that response further and are taken up first.

The substantial gap observed between anti-tank and anti-personnel average precision replicates one of the most stable findings in this literature and is not specific to synthetic training data. Work on the same real dataset [13] reported anti-tank mines at 60.9% mAP@0.5 against 19.3% for anti-personnel mines, and demonstrated that this gap exceeded the differences between architectures, concluding that mine size constrains detection more fundamentally than architectural sophistication. Surveys of tiny object detection identify small target extent as a principal failure mode for standard detection architectures [79], [80], and the introduction of focal loss was motivated by the related problem of extreme foreground-to-background imbalance, which is pronounced in aerial landmine imagery where target pixels form a very small fraction of any frame [57].

A resolution confound amplifies this effect, because it is a property of the present generator configuration rather than of synthetic data in general. All synthetic imagery is produced at 256 by 256 pixels, while real AMLID imagery is 640 by 512. Since the detector resizes all inputs to 640, synthetic images are upsampled by a larger factor than real ones, and upsampling cannot restore detail that was never rendered. The effective resolution of a synthetic anti-personnel mine is therefore lower than its pixel dimensions suggest, and the smallest targets suffer disproportionately. Recent fusion work restricting cross-attention to a local pixel neighborhood was motivated by the same observation, that small-object performance is limited by the preservation of local detail [46].

Helpfully, the GAN-Real source provides a clean way to isolate this factor. Those images are real AMLID scenes downscaled to 256 by 256 and restyled, so they carry real-world content at synthetic resolution. If content realism were the dominant constraint, GAN-Real would be expected to outperform Blender renders substantially, since only one of the two carries real scene content. In the substitution mixes, where real imagery anchors training, the two sources do perform within three points of one another at every matched share, which points to resolution rather than content realism as the binding constraint in that regime. At fully synthetic composition, however, the two diverge, with GAN-Real ahead by 9.8 points for YOLOv11 and 18.2 for RF-DETR, and GAN-Real alone degrades gracefully with altitude (Section IV-A), so content realism reasserts itself once no real imagery anchors the mix. This has a direct practical implication: generating synthetic imagery at native sensor resolution is likely to yield a larger improvement than further refinement of appearance at the current resolution, and it is the more economical of the two interventions.

A landmine is also, by design, a concealed target. The camouflaged object detection literature addresses the general problem of objects that blend into their background and has produced architectural components, including edge-aware decoders and texture-difference modules, that are directly applicable to this task [81], [82].

### F. FUSION BEHAVIOR AND SURVEY CONDITIONS (RQ3)

Results across fusion levels agree with the established finding that RGB information dominates in this task and that thermal input functions as a complementary enhancement, not a substitute, but they do not reproduce the interior optimum reported previously, and the difference should be stated. The predecessor study on this dataset [13] found optimal performance at 10 to 30% LWIR contribution, with pure thermal input producing a 36.7% deficit against optimal fusion while pure RGB trailed that optimum by only 0.1 points. In the present results no thermal fraction exceeds the 0% value. Accuracy in the real-only models is flat from 0 through 30% thermal, at 79.4, 78.8, 79.2, and 78.6 mAP@0.5, and declines thereafter to 44.5 at pure LWIR. What reproduces, then, is the tolerance band and the collapse at the thermal extreme, a 34.9-point deficit for pure LWIR against the 36.7 points reported previously. What does not reproduce is any measurable gain from adding thermal signal. Given that pure RGB and the 10 to 30% band differ by less than a point in both studies, the two findings are better read as a shared plateau whose interior peak is within noise than as a contradiction, and the safer claim for this dataset is that thermal contribution up to roughly 30% is free, not beneficial.

That the plateau-and-collapse pattern reproduces under synthetic and refined training data is a meaningful validation independent of the mixing results, and it holds in the stratified analysis of Section IV-C as well as in the real-only curves. It indicates that the generation and refinement pipeline preserved the relative information content of the two channels even where it did not reproduce their absolute appearance, which is a non-trivial property for a pipeline that composites two registered bands by alpha blending rather than simulating a fused sensor. The qualification is that preserving relative information content is a weaker claim than preserving the thermal signature itself: a pipeline that renders thermal contrast approximately, without sensor noise and with only three blanket materials, could plausibly retain the ordering of the fusion curve while understating the absolute value of the thermal channel. The absence of a measurable thermal benefit here is therefore consistent with either a genuine property of the task or a limitation of the generator, and the present design cannot separate the two.

Approaches that avoid committing to a single fixed ratio, including detectors trained on randomly mixed patches of both modalities [45] and detection transformers that fuse adaptively per query [41], represent a different response to the same underlying observation and are noted as directions the fixed-ratio design of this study does not explore.

## G. LIMITATIONS IN RELATION TO PRIOR WORK

Several constraints should be read alongside the agreements described above. The rendering pipeline does not reproduce sensor-level thermal noise, and environmental parameters were not matched to recorded conditions from the real dataset. Physically based thermal simulation using finite-volume heat transfer to compute surface temperatures for infrared rendering [68] indicates the level of modeling required for full physical fidelity, and the present generator does not reach it. The consequence is that the rendering-fidelity comparison conducted here spans a narrower fidelity range than the literature against which it is compared, and conclusions about physical fidelity should be read as applying only to the range actually sampled.

Similarly, the environmental simulation behind the synthetic datasets was more rudimentary than would be ideal, a consequence of the study's timeframe and computational budget rather than of a principled choice. AgX view transforms, soft shadows, and per-window yaw and position randomization were sufficient to bring the renders within useful distance of real captures, but physically based ray tracing, high-dynamic-range lighting, measured surface BSDFs, and higher-resolution terrain, vegetation, and surface weathering would each have raised fidelity further. The effect of that additional fidelity on downstream detection remains untested here.

In addition, many of the landmine models were built from basic geometry with simplified external casings that captured only the general silhouette and major features of each device. This may be adequate for higher capture altitudes, where a mine occupies only a few pixels, but the simplification becomes conspicuous at low altitude, where the absence of finer geometric detail such as surface texture and relief is apparent and may contribute to the accuracy differences observed between models trained on real and on synthetic imagery. The material treatment compounds this. As stated in Section III-B, each mine carries one of three blanket materials with a uniform set of thermal constants, whereas real mines exhibit paint wear, oxidation, adhering debris, and surface roughness that would break up the thermal signature rather than presenting it uniformly.

Environmental complexity was constrained by the same budget. The scene uses a single, largely flat ground surface of blended grass and dirt, without the rocks, debris, or man-made clutter a real demining site would present, and without significant slopes, depressions, or areas of disturbed soil. A broader range of ground types spanning different soil densities and moisture contents would introduce greater variation in both visual appearance and thermal response, and would plausibly improve the transferability of models trained on this imagery.

A further constraint bears directly on the comparison between synthetic and real thermal imagery. A rigorous comparison would require that synthetic frames be generated under the same environmental conditions as the real captures they are measured against, but the AMLID dataset does not carry the recorded environmental metadata, including ambient temperature and the conditions prevailing at capture, that such matching would require. The limitation is therefore not that these parameters were left unmatched by choice, but that the information needed to match them was unavailable.

Furthermore, the two-class label mapping, adopted because the synthetic annotations do not distinguish casing material, reduces the granularity of the comparison relative to the four-category taxonomy available in the real dataset. This is a material loss, because the predecessor study found that casing material produced a 14.4% difference within anti-tank mines, attributable to the differing thermal conductivity of metal and plastic [13]. The present design cannot observe that effect, and cannot determine whether synthetic data reproduces it.

Likewise, the resolution mismatch discussed in Section V-E is a property of this generator configuration rather than of synthetic data in general, and results on the anti-personnel class in particular should not be generalized beyond it. Finally, the refinement network was trained on a single GPU for a fixed iteration budget with checkpoint selection by Earth Mover's distance, a distributional criterion that Section V-B argues is an imperfect proxy for training utility. Selecting the refinement checkpoint by downstream detection performance rather than by distributional similarity is a straightforward extension that the present design does not implement.

## VI. CONCLUSION

This study asked how far rendered and refined imagery can substitute for real multispectral imagery in landmine detection, and the answer is bounded but useful. Across size-matched training sets of 10,756 images, replacing up to half of the real imagery with Blender renders cost under 2.8 mAP@0.5 points against real-only baselines of 73.1 for RF-DETR and 71.4 for YOLOv11, with the strongest substituted configurations recovering 97.7 and 98.0% of those baselines. Beyond a 70% synthetic share the losses grow by roughly an order of magnitude, and training entirely on rendered imagery retains only half of the RF-DETR baseline and a third of the YOLOv11 baseline. Augmentation behaves differently from substitution. Pooling every available source into a set four times larger raised YOLOv11 to 76.3, a gain of 4.92 points and the only configuration in the study to exceed a real-only baseline, while the same pooling cost RF-DETR 3.33 points. Notably, refinement did not help where it was expected to: applying the CycleGAN refiner to Blender renders produced the two weakest sources in the study, at 23.0 and 14.1 mAP@0.5, consistent with the two generative stages that separate that imagery from the real sensor domain.

Where synthetic imagery contributes depends on the conditions under which the real imagery was acquired, and substitution and augmentation pull in opposite directions along the same axis. The penalty for substituting renders is negligible where the fused input is RGB-dominant and grows with thermal share, reaching 4.1 points for RF-DETR at pure LWIR and 6.3 for YOLOv11 at a 90% thermal share, and 12.3 points for YOLOv11 at a 70% substituted share. Augmentation inverts that pattern, improving YOLOv11 by 11.8 points at pure LWIR against roughly 3 to 4 points in the RGB-dominant band, and by 6.7 points at 10 m against 2.7 at 5 m. Synthetic imagery therefore cannot replace real thermal imagery but can reinforce it, and the practical rule that follows is to substitute renders only within the RGB-dominant envelope and to retain them as an additive supplement in the thermal-dominant conditions where real data is weakest. Altitude and season are comparatively weak moderators, and the time-of-day effect reverses between seasons, so neither supports an acquisition recommendation on its own.

For a demining program deciding between investment in field collection and investment in rendering capacity, the practical reading is that rendering can absorb a substantial share of the data burden but cannot carry it alone. Half of a labeled real corpus can be displaced at a cost small enough to be operationally irrelevant, which matters where survey access is restricted and controlled collection is expensive and hazardous. What rendering cannot do at present is substitute for real imagery entirely, and it cannot rescue the anti-personnel class, whose average precision falls from 48.3 points under real training to between 4.7 and 11.4 under fully synthetic training, while anti-tank precision falls from 92.8 to between 32.4 and 71.2. A program adopting synthetic data should therefore budget for real collection sufficient to hold the smallest target class, and should treat rendering as a multiplier on that corpus, not a replacement for it. The corollary for tooling is that adversarial refinement of rendered input is not a safe default, since it degraded every configuration in which it was stacked on renders.

Two extensions follow directly from the design and are inexpensive relative to their value. The first is to sample real-data fractions between 0 and 30%, which is where the present ladder is silent and where the knee in the response curve must lie; the current sampling can establish that a plateau exists but not where it ends. The second is to generate synthetic imagery at native sensor resolution rather than at 256 by 256. Because the detector resizes all inputs to 640, synthetic frames are upsampled by a larger factor than real ones, and upsampling cannot restore detail that was never rendered, which imposes a ceiling on the smallest targets in particular. The refined-real source carries real scene content at synthetic resolution, yet in the substitution mixes it performs within three points of Blender renders instead of substantially better, which points to resolution, not appearance, as the binding constraint wherever real imagery anchors training, and makes native-resolution generation the more economical of the two interventions. Only at fully synthetic composition, where GAN-Real leads every other source, does content realism reassert itself.

## ACKNOWLEDGMENTS

This research was made possible through the support of George Mason University's College of Science, which supports the Aspiring Scientists Summer Internship Program (ASSIP).

Portions of this research used generative AI. Claude Opus 5 vision-language model assisted in constructing the Blender scenes through a Model Context Protocol server, as described in Section III-A; it did not generate or modify any annotation. AI assistance was also used in developing analysis code, preparing figures, and editing the manuscript. All AI-assisted output was reviewed and verified by the authors, who take full responsibility for the content of this article.

## AUTHOR BIOGRAPHIES

EVAN LI is a second-year student at Thomas Jefferson High School for Science and Technology. He is a senior member of a globally recognized VEX VRC team, having advanced to Division Finals at the VEX VRC World Championship, and he helped qualify TJHSST for the NASA Student Launch Initiative through his work in the American Rocketry Challenge. His interest in autonomous systems and sensor-driven robotics has led him to explore synthetic imagery generation for multispectral UAS-based detection.

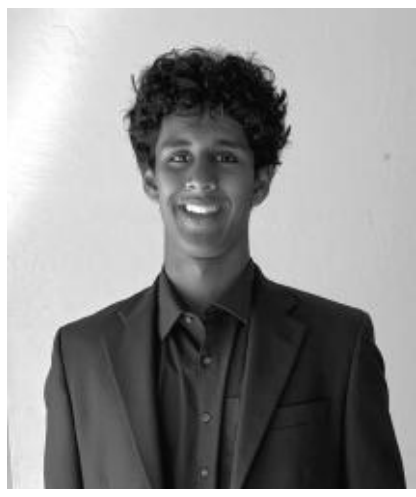

BHARGAV MANDAKOLATHUR is a sophomore at Milpitas High School in Milpitas, California. He has received the Gold Presidential Volunteer Service Award for three consecutive years for his volunteer commitment. He is interested in applying robotics and AI to real-world problems, and has earned honorable mentions at the national 3M Young Scientist Challenge and the Synopsys Science and Technology Championship for his work in computer vision and thermodynamics. This research was conducted through George Mason University's Aspiring Scientist Summer Internship Program (ASSIP).

KABIR RAGHAVAN is a junior at Walt Whitman High School, Bethesda, MD, and a member of the class of 2028. He has competed in the FIRST Tech Challenge and the FIRST Robotics Competition and has conducted research on humanoid robots in Mountain View, CA. He teaches astronomy and mathematics to middle school-aged students in rural Tamil Nadu, India, and his current research uses an unmanned aerial platform to consider different facts affecting air quality.

ARYAN RUIA is currently pursuing a high school diploma from the Thomas Jefferson High School for Science and Technology, where he is the Head Lecturer of the Quantum Physics Club, and President of the Quantum Computing Club. He is a member of low and high power rocketry teams, participating in the American Rocketry Challenge and the NASA Student Launch Initiative.

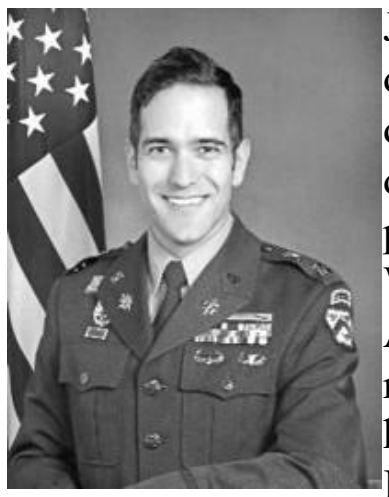

JAMES E. GALLAGHER received his M.S. and Ph.D. degrees from George Mason University, where his doctoral research focused on multispectral object detection for drone-based applications. He is currently pursuing a Ph.D. in systems engineering at George Washington University. He is also an active-duty U.S. Army Major in the Military Intelligence Corps. His research in RGB-LWIR object detection was inspired by his experience using multispectral sensors from UAS in Iraq.

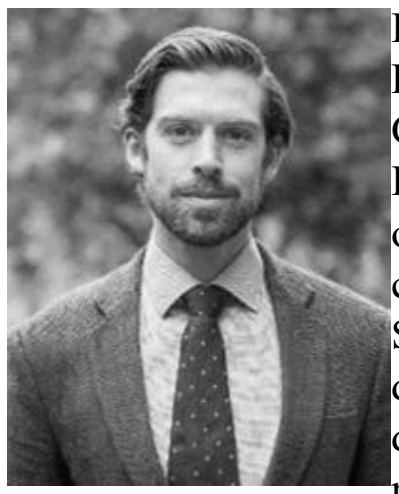

EDWARD J. OUGHTON received the M.Phil. and Ph.D. degrees from Clare College, University of Cambridge, U.K., in 2010 and 2015, respec-tively. Later, he held research positions at both the University of Cambridge and the University of Oxford. He is currently an Assistant Profes-sor with the College of Science, George Mason University, Fairfax, VA, USA, developing open-source research software to analyze digital infras-tructure deployment strategies. He received the Pacific Telecommunication Council Young Scholars Award, in 2019; the Best Paper Award from the Society of Risk Analysis, in 2019; and the TPRC48 Charles Benton Early Career Scholar Award, in 2021.